\RequirePackage{snapshot} 
\documentclass{aiaa-tc} 

\usepackage{graphicx}
\usepackage[space]{grffile}
\usepackage{latexsym}
\usepackage{enumerate}
\usepackage{textcomp}
\usepackage{longtable}
\usepackage{multirow,booktabs}
\usepackage{amsfonts,amsmath,amssymb}
\DeclareMathOperator*{\argmax}{\arg\!\max}
\usepackage{nomencl}
  \makenomenclature
\usepackage{url}
\usepackage{algorithm}
\usepackage{algpseudocode}
\usepackage{hyperref}
\hypersetup{colorlinks=false,pdfborder={0 0 0}}
\usepackage[utf8]{inputenc}
\usepackage{bibentry}
\title{Towards Adaptive Training of Agent-based Sparring Partners for Fighter Pilots}

\author{%
    Brett W. Israelsen\thanks{Graduate Researcher, Computer Science, University of Colorado, Boulder, CO, AIAA Student Member}
    \ and
    Nisar Ahmed\thanks{Assistant Professor, Aerospace Engineering Sciences, University of Colorado, Boulder, CO, AIAA Member}\\
    {\normalsize\itshape
    University of Colorado, Boulder, Colorado, 80309, USA}\\
    Kenneth Center\thanks{Director, Orbit Logic Incorporated, Greenbelt, MD, AIAA Professional Member}
    \ and
    Roderick Green\thanks{Sr. Software Engineer, Orbit Logic Incorporated, Greenbelt, MD} \\
    {\normalsize\itshape
    Orbit Logic Incorporated, Greenbelt, Maryland, 20770, USA}\\
    Winston Bennett Jr. \thanks{Division Technical Advisor -- Air Force Research Laboratory Warfighter Readiness Research Division, Wright Patterson AFB, Ohio} \\
    {\normalsize\itshape
Wright Patterson AFB, Ohio, 45433, USA}
 }

 \AIAApapernumber{200?-????}
 \AIAAconference{InfoTech, 8-13 January 2017, Grapevine Texas}
 \AIAAcopyright{\AIAAcopyrightD{200?}}

\newcommand{\BO}{GPBO}

\begin{document}
\maketitle

\begin{abstract}
    A key requirement for the current generation of artificial decision-makers is that they should adapt well to changes in unexpected situations.  This paper addresses the situation in which an AI for aerial dog fighting, with tunable parameters that govern its behavior, must optimize behavior with respect to an objective function that is evaluated and learned through simulations. Bayesian optimization with a Gaussian Process surrogate is used as the method for investigating the objective function. One key benefit is that during optimization, the Gaussian Process learns a global estimate of the true objective function, with predicted outcomes and a statistical measure of confidence in areas that haven't been investigated yet. Having a model of the objective function is important for being able to understand possible outcomes in the decision space; for example this is crucial for training and providing feedback to human pilots. However, standard Bayesian optimization does not perform consistently or provide an accurate Gaussian Process surrogate function for highly volatile objective functions. We treat these problems by introducing a novel sampling technique called Hybrid Repeat/Multi-point Sampling. This technique gives the AI ability to learn optimum behaviors in a highly uncertain environment. More importantly, it not only improves the reliability of the optimization, but also creates a better model of the entire objective surface. With this improved model the agent is equipped to more accurately/efficiently predict performance in unexplored scenarios.    
\end{abstract}

\section{Introduction}
    Rapid advancement of capabilities in the Artificial Intelligence (AI) research area have the potential to completely revolutionize the way that U.S. armed forces train for battlespace dominance. As AI becomes more sophisticated, there are many opportunities to insert it into domain training environments.

    One use of AI is as the basis of agents that stand in for humans in Live-Virtual-Constructive (LVC) simulations. Humans participating in training exercises in these environments need to be challenged at an appropriate level relative to their skills. Red-force agents that can assess the skill-level of human participants and adapt accordingly are needed. Agents like this could then serve as worthy and credible sparring partners that are indistinguishable from the most experienced humans.

    Specifically, in this work we investigate how an AI agent with tunable parameters that govern overall behavior can be adapted to optimize an objective function for engagement outcomes. Beyond optimizing some outcome metric, it is also important that the agent have a realistic representation of the entire objective function to enable the AI to adapt its behavior appropriately in dynamic and uncertain environments. These behavioral changes could be based on adapting to the adversaries skill level (it is not desirable to use the same difficulty level for novice and advanced users), or the user's ability to exploit a weakness of the agent (the objective function is changing).

    There are several challenges that make optimization of decision-making AI difficult in this application:
    
    \begin{enumerate}
        \item Simulating an engagement can be costly. Beyond the financial expense of operating the simulation environment, contributions to the cost may also include the involvement of skilled personnel with limited availability, and the wall-clock duration of the simulation itself.\label{pnt:cost}
        \item The engagement metrics that need to be optimized cannot be described analytically, but can only be sampled by running simulations. When sampled they are generally nonlinear and noisy. Consequently, many traditional optimization methods are not applicable.\label{pnt:uglyobjective}
        \item Due to the realistic nature of the simulations and the nature of aerial combat, the objective function is extremely volatile and uncertain (e.g. due to combined random effects of weather, nonlinear gust modeling for wind, terrain, sensor noise, psycho-motor time delays, etc.).\label{pnt:volatile}
        \item Besides only identifying the optimum performance, the agent should also try to obtain some model of the overall objective function. This can allow the agent to be adaptive and have some notion of what outcomes might arise when modifying behavior parameters, without having to exhaustively search over a high-dimensional parameter space. In addition, this model can also be used to generate a useful estimate of the expected performance of adversaries for a wide range of scenarios, using only a small number of test evaluations. \label{pnt:objectivemodel}
    \end{enumerate}

    Bayesian optimization using Gaussian Process surrogate functions (abbreviated as \BO) is well-suited for addressing points \ref{pnt:cost} and \ref{pnt:uglyobjective}. \BO{} uses the Gaussian Process (GP) surrogate function to approximate the true objective function. This surrogate function includes an estimated value of the objective function at some location in the solution space as well as the confidence in that estimated value. The optimization algorithm then makes use of this surrogate function to intelligently search the solution space for the optimum, based on a number of sampled function evaluations, i.e. using `explore/exploit' strategies to locate the minimum as quickly as possible while using sparse function evaluations to build the surrogate model.

    We show that standard application of \BO{} methods are not well suited to address points \ref{pnt:volatile} and \ref{pnt:objectivemodel}, in this application. We demonstrate a novel approach to implementing \BO, called Hybrid Repeat/Multi-point Sampling (HRMS), that addresses this issue. In this setting, \BO{} with HRMS is able to not only identify the minimum more reliably than standard \BO, but also yields a more useful surrogate representation of the objective surface. Finally, it does this using no more total function evaluations than traditional \BO.

    In the remainder of the paper some of the previous work that has occurred in this field is discussed, and a formal definition of the problem is given, including details regarding the application of \BO{} for aerial dog fighting. Section \ref{sec:background} reviews some of the previous work in this area, and gives some theoretical background for \BO. A formal definition of the adaptive agent problem, and the `learning engine' framework that is used to train the decision-making AI is given in Section \ref{sec:methodology}.  Finally, in section \ref{sec:results} we show by simulated experiments, that our implementation of \BO{} with HRMS is useful for optimizing highly volatile metrics for this application, as well as different AI parameters. Furthermore, HRMS outperforms traditional sampling techniques, thus yielding valuable insights about fighter performance through the global surrogate function GP model.

\section{Background}\label{sec:background}
    Simulations for air-to-air combat training have been studied extensively. In their 1990 paper, McManus and Goodrich discuss the integration of an AI system into two separate simulators, one of which included an interface for pilots to participate~\cite{McManus1990}. Moore et. al discussed a method to measure pilot performance and validated it in simulated combat scenarios~\cite{Moore1979}.

    In order to evaluate a pilot (human or AI) performance metrics must be calculated. Traditionally performance grades have been  issued by flight instructors, but have evolved to be less subjective and more repeatable. Many of these metrics are well accepted in the community. In contemporary development of newer metrics, expert evaluations are still utilized for validation. In his paper, Kelly reviews and summarizes much of this work~\cite{Kelly1988}. He specifically mentions metrics that include relative aircraft position, throttle and speedbrake manipulation, and overall engagement outcomes to name a few. Producing meaningful metrics that operate on time-series and summary data from engagements is still an active field of research~\cite{Kelly1988,Mulgund1998,Huynh1987,McManus1990,Mulgund2001,Paranjape2006}. Our experiments utilized some of the metrics mentioned in the literature above. In this paper we focus on the `cumulative time to kill' (cTTK) metric, which is total time to eliminate the enemy.

    The current body of work that involves optimization in aircraft engagement is focused mainly on optimal strategies of teams. Mulgund et. al were concerned with ``large-scale'' air combat tactics (formations, etc.) and were able to demonstrate promising results in that area~\cite{Mulgund1998,Mulgund2001}. Wu et. al addressed the problem of optimizing cooperative multiple target attack using genetic algorithms (GAs)~\cite{Wu2005}. Also applying GAs, Gonsalves and Burge investigated how mission plans could be optimized~\cite{Gonsalves2004}. This optimization is focused on the strategic level, whereas this work is focused on modifying behaviors of AI pilots.
    
    Recently there has been work by Ernest et. al to design an AI that can function in real time during combat while also being interpretable~\cite{Ernest2016}. Their work utilized genetic optimization of a group of Fuzzy Inference Systems (FISs) that together produce control actions given input data. This system is able to produce `fine-grained' control actions of a single jet or groups of fighters. Also, due to the specification of the FISs the control outputs are interpretable to humans. This work is fairly similar to ours in that it seeks an optimal solution to a combat scenario by modifying fighter behaviors, while also being interpretable. Their solution differs in that they design the entire AI while our work operates on top of any AI with tunable behaviors. Also, our approach tries to minimize the number of simulations needed to find the optimum, and build a model of the solution space at the same time; this model can then be used later by the AI to adapt decisions.

    The use of autonomous agents in simulation scenarios is used extensively in the video game industry where a non-player character (NPC) is the autonomous agent that interacts with human players. The majority of agents used in the video gaming industry are using 30 year old technology that is highly scripted~\cite{Yannakakis2012}. Cole et. al used GAs in order to tune game agents for first person shooter games~\cite{Cole2004a}. Liaw et. al used genetic algorithms in order to evolve game agents that work as a team~\cite{Liaw2013}. Othman et al discuss the use of simulations to evolve an AI agent for tactical purposes~\cite{Othman2012a}. This work addresses the optimization of behaviors, but does not take into account the cost of running experiments or constructing a model of the true objective function to allow the AI to be more robust given an extremely volatile objective metric.

    Finally, Bayesian optimization has emerged as a critical tool for tuning hyperparameters in various areas of machine learning ~\cite{Bergstra2011,Snoek2012,Mahendran2012}. It has also been applied in several other fields such as modeling of user preferences, and reinforcement learning~\cite{Brochu2010}. It is well suited for optimizing objective functions that are expensive to evaluate and unknown. It often does this with the fewest function evaluations as compared to other competing methods~\cite{Jones1998}. These properties make it ideal for optimizing the behavioral parameters of an AI pilot. However, as already mentioned and as will be demonstrated later, standard application of \BO{} is not able to perform well on objective functions that are highly volatile, or when a surrogate representation of the true objective function is desired.

    \subsection{Bayesian Optimization}\label{bayesopt}
    The goal of any optimization is to minimize some nonlinear function $y:X \rightarrow \mathbb{R}^d$ \nomenclature{$f$}{nonlinear objective function} \nomenclature{$X$}{optimization search space, or solution space} \nomenclature{$\mathbb{R}^d$}{d-dimensional Euclidean space} \nomenclature{$d$}{dimension of the search space} ($y$  is a function that maps from $X$  to $\mathbb{R}^d$). Here, $X$ is typically a subset of Euclidean space $X \in \mathbb{R}^d$ known as the search space, or solution space. The function $y$ is known as the objective function. The task of minimization is then to search over $X$ to identify an element $x^* \in X$ \nomenclature{$x^*$}{optimum in $X$} such that $f(x^*) \leq f(x)\forall x \in X$ (i.e. the function at $x^*$ is less than the function for all other values of $x$ in $X$).

    In classical optimization theory one assumes the mathematical description of $y$ is known (i.e. $y(x) = x^{2} + 3x^{3}$). When $y$ is known it is fairly easy (quick) to evaluate. In many applications (like the one in this paper) this is often not the case. Specifically, the only knowledge of $y$ might be noisy function evaluations.

    The goal of Bayesian optimization is to minimize some unknown, noisy objective function that is costly to evaluate, while also learning about it at the same time. It is so named because Bayesian inference, a key concept from  probability theory, is applied during the optimization process. That is, that there is an initial belief over potential objective functions $p(y)$, called a prior, that can be updated by subsequent observations or evidence $p(Z(x)|y)$. Mathematically this leads to an application of Bayes' rule: $p(y|Z(x)) \propto p(Z(x)|y)p(y)$ (see appendix A.1 of Rasmussen's book for more detail~\cite{Rasmussen2006}). The quantity $p(Z(x)|y)$ is also known as the likelihood. Finally, $p(y|Z(x))$ is the updated probability of $y$ given that $Z(x)$ has been observed, also known as the posterior. In words, given some prior belief about the shape of the noisy objective function and evidence about the shape of the function, an updated posterior belief about $y$ can be formed. In the limit as the number of observations approaches infinity, the posterior $p(y|Z(x))$ will approach the true underlying $y$, due to the law of large numbers.
    \begin{figure}[htbp]
        \centering
        \includegraphics[width=0.90\textwidth]{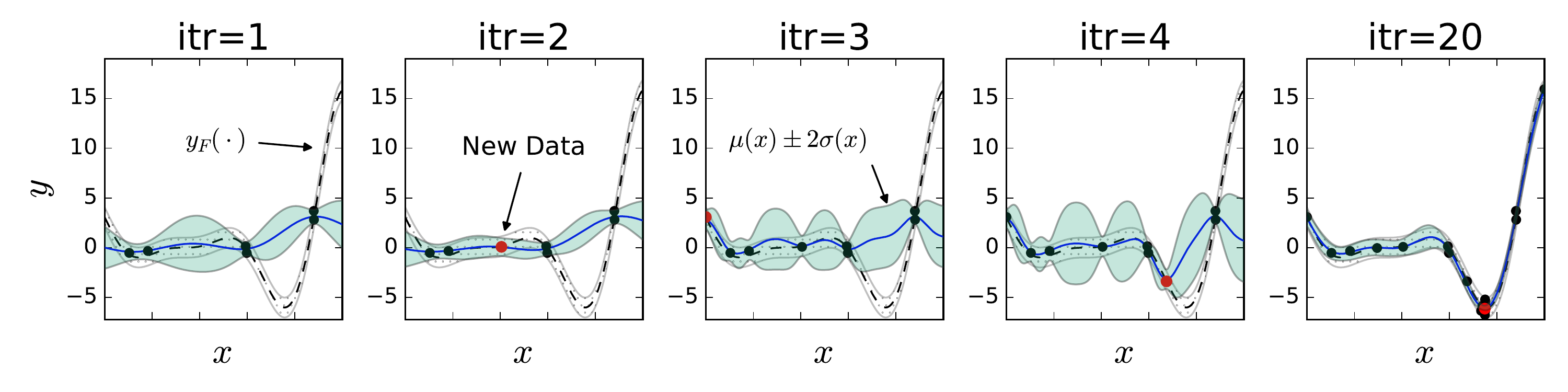}
        \caption{Simple example of \BO{} progression. Begin with some seed data and fit a GP to it, add data at `interesting' location as quantified by the acquisition function (defined later in this section). Add new data to the GP and continue.}
        \label{fig:1dexample}
    \end{figure} 

    Concretely, the main components of the Bayesian optimization process are:

    \begin{enumerate}[(a)]
        \item The surrogate function, which represents the initial belief over potential objective functions
        \item The acquisition function, which provides a way for the algorithm to ``intelligently'' search the solution space $X$ to find the optimum.
    \end{enumerate}

    A simple example of the \BO{} progression on a 1d objective function is shown in Figure \ref{fig:1dexample}. The function $y_{F}(\cdot)$ is known as the Forrester function and is a standard objective function for testing optimization methods~\cite{forrester2008engineering}. Iteration 1 begins with some seed data to fit an initial surrogate function. The GP surrogate function is represented by the blue line, and shaded confidence bounds. The acquisition function (not shown, but defined later in this section) directs where new data is to be sampled. When new data is acquired the inference along $x$ is updated, and the process continues. In this case at iteration 20 (the far right column) the minimum has been located and the GP represents the true objective function fairly well.

    \paragraph{The Surrogate Function: }\label{sec:gaussian-processes}

    Gaussian processes (GPs) are defined as a probability distribution over functions $y(x)$ such that the set of values of $y(x)$ evaluated at an arbitrary set of points, jointly have a Gaussian distribution~\cite{Bishop2006,Rasmussen2006}. When used in Bayesian optimization, a Gaussian process can serve as a `surrogate' function, or an estimate of the true objective function. This is referred to as Bayesian optimization with a Gaussian Process surrogate (GPBO). There are two main components of a Gaussian process shown in Equation \ref{eq:gp}: The mean function (equation~\ref{eq:gp_mean}), and the covariance function (equation~\ref{eq:gp_cov}). In theory $m(x)$ could be any function but we will assume $m(x)$ is zero in this paper to simplify notation.
    \begin{align}
        m(x) &= \mathbb{E}[f(x)] \label{eq:gp_mean} \\
        k(x,x^\prime) &= \mathbb{E}[(f(x)-m(x))(f(x^\prime)-m(x^\prime))] \label{eq:gp_cov} \\
        f(x) &\sim \mathcal{GP}(m(x),k(x,x^\prime)) \label{eq:gp}
    \end{align}

    The covariance function (which is also referred to as the covariance kernel) used in this work is the Mat\'{e}rn kernel.
    \begin{align}
        k_{\nu = 3/2}\left( r \right) &= \left( 1 + \frac{\sqrt{3}r}{l} \right)\exp\left( - \frac{\sqrt{3}r}{l} \right) \label{eq:Mat32}
        \nomenclature{$\nu$}{differentiability parameter of Mat\'{e}rn kernel}
        \nomenclature{$l$}{Kernel hyperparameter -- length scale}
    \end{align}

     The kernel is commonly used in machine learning applications because of the property of being able to change the smoothness by using the parameter $\nu$~\cite{Rasmussen2006}. The parameter $\nu=3/2$ means that the resulting function will be differentiable once. The quantity $r = |x - x^\prime|$, and $l$ is a `length scale' parameter.
     
     Given $n$ training observations, the individual elements of the covariance kernel make the covariance matrix $K(X,X)\in \mathcal{R}^{n\times n}$ with elements $k(x_i,x_j)$ that are the covariance between $x_i$ and $x_j$ for all pairs of training data. If there is a Gaussian likelihood of the data given the observations, then the joint distribution of the $n$ training outputs and $p$ test outputs $f(X)\in\mathcal{R}^{n\times 1}$ and $f(X_*)\in\mathcal{R}^{p\times 1}$ respectively can be written as shown in equation~\ref{eq:fX}. Thus, given the previously measured values $X$ and $f(X)$ predictions of $f_*(X_*)$ can be made.
    \begin{align}
        \begin{bmatrix}
            f(X) \\
            f_*(X_*) \end{bmatrix} &= \mathcal{N}\left(\textbf{0},\begin{bmatrix}
                K(X,X) & K(X,X_*) \\
                K(X_*,X) & K(X_*,X_*)\end{bmatrix}\right) \label{eq:fX}
    \end{align}

    For inference, the conditional values of the test outputs are of interest:
    \begin{align}
        f_{*}|X_{*},X,f &\sim \mathcal{N}(\mu(X_*),\sigma^2(X_*)) \label{eq:predict2} \\
        \mu(X_*) &= K\left( X_{*},X \right)K\left( X,X \right)^{- 1}f \label{eq:mu}\\
        \sigma^2(X_*) &= K\left( X_{*},X_{*} \right) - K\left( X_{*},X \right)K\left( X,X \right)^{- 1}K\left( X,X_{*} \right) \label{eq:sigma}
    \end{align}
    \nomenclature{$\mu$}{Conditional mean of $f_*$}
    \nomenclature{$\sigma^2$}{Conditional variance of $f_*$}
    Here, $K(X,X_*)\in \mathcal{R}^{p \times n}$ making $\mu(X_*)\in\mathcal{R}^{p\times 1}$ and $\sigma^2(X_*)\in\mathcal{R}^{p \times p}$. Equation \ref{eq:predict2} shows the expression of the conditional distribution $f_*$ given test points $X_*$, and training data $X$ and $f$. The mean and variance of that distribution can be calculated analytically using equations \ref{eq:mu} where and \ref{eq:sigma}.

\paragraph{The Acquisition Function: }\label{acq_fxn}

    The acquisition function, $a(\cdot)$, determines the location of the next function evaluation or experiment. It typically operates on the surrogate function to quantify the possibility of finding the optimum of the objective function at some location. An acquisition function indicates where the optimum is most likely to be found, according to the current information about the objective function. Bayesian optimization selects the $x$ at the optimum of the acquisition function as the next point to be evaluated. In this paper three acquisition functions will be evaluated, they are: Expected Improvement (EI), GP Upper Confidence Bound (UCB), and Thompson Sampling (TS).
    
    Acquisition functions, in themselves, have many local optima and require global optimization methods to identify the optimum. It is, therefore, necessary to use global optimization algorithms. Recall that the only reason this makes sense is because the true objective function is assumed to be `expensive' to evaluate. Because of that expense, it is not acceptable/desirable to forego using \BO{} and try to use a more standard global optimization method. In other words, global optimization of the acquisition function is a small cost to pay when the true objective function is expensive enough.
    
    This work makes use of the DIRECT (dividing rectangles) optimization algorithm that utilizes the Lipschitz continuity property to bound function values in local rectangles and search accordingly~\cite{Jones1993}. DIRECT is also commonly used in \BO{} (see \cite{Ponweiser2008,Hoffman2011,Mahendran2012,Wang2013}). More discussion on this topic can be found in in some work by Shahriari et al. \cite{Shahriari2015a}.
    
    \subsubsection{Expected Improvement:} The EI function is defined by~\cite{Jones1998}
    \begin{align}
        \text{EI}(f(x)) &= \begin{cases}
                            (\mu(x) - f(x^{*}))\Phi(Z) + \sigma(x)\phi(Z) &,  \quad \sigma(x)>0 \\
                            0 &, \quad \sigma(x)=0
                        \end{cases} \label{eq:EI1}\\
        Z &= \frac{\mu\left( x \right) - f(x^{*})}{\sigma(x)} \label{eq:EI2}
        \nomenclature{$\mu(x)$}{predicted mean at $x$}
        \nomenclature{$x^*$}{optimum observed value}
        \nomenclature{$\sigma(x)$}{predicted standard deviation at $x$}
        \nomenclature{$Z$}{Standard normal distribution}
        \nomenclature{$\Phi$}{PDF -- probability distribution function of the standard Normal distribution}
        \nomenclature{$\phi$}{CDF -- cumulative distribution function of the standard Normal distribution}
    \end{align}
    Here, $\mu(x)$ is the mean predicted value of the GP at $x$, and $\sigma(x)$ is the predicted standard deviation at $x$. $f(x^{*})$ is the optimum observed value of the objective function\footnote{ This should be substituted by the mean predicted optimum if the objective function is stochastic}. Also, $\Phi(\cdot)$ and $\phi(\cdot)$ represent the PDF and CDF of the standard normal distribution.

    What makes EI unique is that it explicitly uses the current observed optimum of the objective function along with the predicted $\mu$ and $\sigma$ to quantify where the optimum might be located. As the name suggests, the values of EI literally translate to the statistically expected improvement in the current estimated optimum when sampling at a given location.
    
    \subsubsection{Upper Confidence Bound:} The UCB acquisition function was inspired by the UCB regret in multi-armed bandit problems~\cite{Srinivas2010}. The key goal is to bound regret in a sequential optimization process, where regret is defined as the difference between the actual strategy (based on decisions made with imperfect information) and the ideal strategy. UCB is defined as
    \begin{align}
        \text{UCB}(x) &= \mu(x)+\beta \sigma(x) \label{eq:UCB}
    \end{align}

    The parameters $\mu$ and $\sigma$ are from the GP, and $\beta$ is a hyperparameter. UCB simply takes the mean prediction and some multiple of the standard deviation at every $x$ and adds them\footnote{In a minimization problem we would use a lower confidence bound by subtracting $\sigma$ in equation \ref{eq:UCB}}. One possible drawback for this function is that it requires a hyperparameter $\beta$. While there exist principled ways to select it~\cite{Shahriari2015a}, UCB is often avoided in practice because of this tuning problem, and the fact that it has not proven to be consistently better than other methods (such as EI) that do not require hyperparameters\cite{Snoek2012}. In our experience selecting a $\beta$ that works for optimization hasn't been very difficult.

    \subsubsection{Thompson Sampling:} Thompson sampling works a bit differently than either EI or UCB. As its name suggests, it involves sampling functions directly from the Gaussian process (using Equation \ref{eq:fX}, this actually means drawing values from discretely sampled locations). Once a function is obtained, the optimum is found and the objective function is evaluated at that point. Exploration and exploitation are inherently part of TS, due to its stochastic nature. Another nice property of Thompson sampling is that it naturally lends itself to parallelized search for the optimum. By the law of large numbers TS will tend to identify locations where there is a better chance of finding the optimum of the objective function.

    It is a known limitation that TS has difficulty performing in high dimensional applications~\cite{Shahriari2015a}. This was also found to be the case for the 1-on-1 aerial combat application at dimensionalities of $d\geq6$ decision making parameters.

    \paragraph{Summary:} Each of the aforementioned acquisition functions has benefits and drawbacks, and, as investigated by Shahriari et al. there is no single best acquisition function for every situation (a manifestation of the `no-free-lunch' theorem) and so we evaluate EI, UCB, and TS in this application~\cite{Shahriari2015a}. We apply each technique experimentally to attempt to identify the best one for the simulated aerial combat scenario.

    Given the functions defined above, the procedure in Algorithm~\ref{alg:BayesOpt} can be followed. The termination criteria could be the number of iterations or something more sophisticated~\cite{Johnson}. In practice the performance of Bayesian optimization depends greatly on the selection and tuning of the surrogate function. We will discuss this further in section~\ref{sec:methodology}~\ref{practical-application-of-bo}.
    \begin{algorithm}[H]
        \caption{BayesOpt()}
        \label{alg:BayesOpt}
        \begin{algorithmic}[1] 
                \While {termination criteria not met}
                    \State $x_t = \argmax_{X} a(f(x))$ \Comment see equation~\ref{eq:UCB}
                    \State Evaluate $f(x_t)$
                    \State Add $f(x_t)$ and $x_t$ to the surrogate function $f(X)$ and update \Comment see equation~\ref{eq:fX}
                \EndWhile
        \end{algorithmic}
    \end{algorithm}

    \subsection{Contributions of this work}
    In this paper we demonstrate the first application of \BO{} to the training of AI decision-makers with continuous behavioral parameters. We show that \BO{} can be used in making AI that is adaptive to uncertain environments with volatile outcomes. We demonstrate that standard application of \BO{} methods are not well suited to addressing highly volatile objective functions like those found in the aerial dog fighting application. These methods also lack the ability to model the overall surrogate function so that it can be used for further decision making and performance analysis.

    We introduce HRMS as a novel sampling approach that addresses the stated shortcomings of standard \BO. By experiments we show that, in the simulated aerial combat scenario, this method not only finds the optimum objective value more reliably/repeatably than standard \BO{}, but it also yields a more accurate surrogate representation of the true objective surface. 

\section{Methodology}\label{sec:methodology}
    The problem is defined as an air combat scenario with autonomous red and blue force agents. Each of the agents has behavioral parameters given by the parameter vectors $\pmb{x}_{r}$ and $\pmb{x}_{b}$ respectively. The goal is to optimize an objective function $y_i(\pmb{x}_{r},\pmb{x}_{b})$, where $y_i(\cdot)$ must be evaluated using a high-fidelity combat simulation. For this work $\pmb{x}_r$ is constant, and the optimization will only be changing $\pmb{x}_b$. The remainder of this paper uses the following: $y=y_{TTK}(\cdot)$ (TTK stands for time to kill), and $\pmb{x_b} = \{x_1,x_2\} = \{intspeed,launch\}$, where $intspeed$ is the intercept speed, (i.e. the speed at which the blue fighter approaches in order to close the range on red once engaged) and $launch$ is the delay in time between lock and launching a weapon (note that there are 11 total behavior parameters available in the aerial-combat simulation, but we only investigate two here).

\begin{figure}
        \centering
        \includegraphics[height=7.0cm]{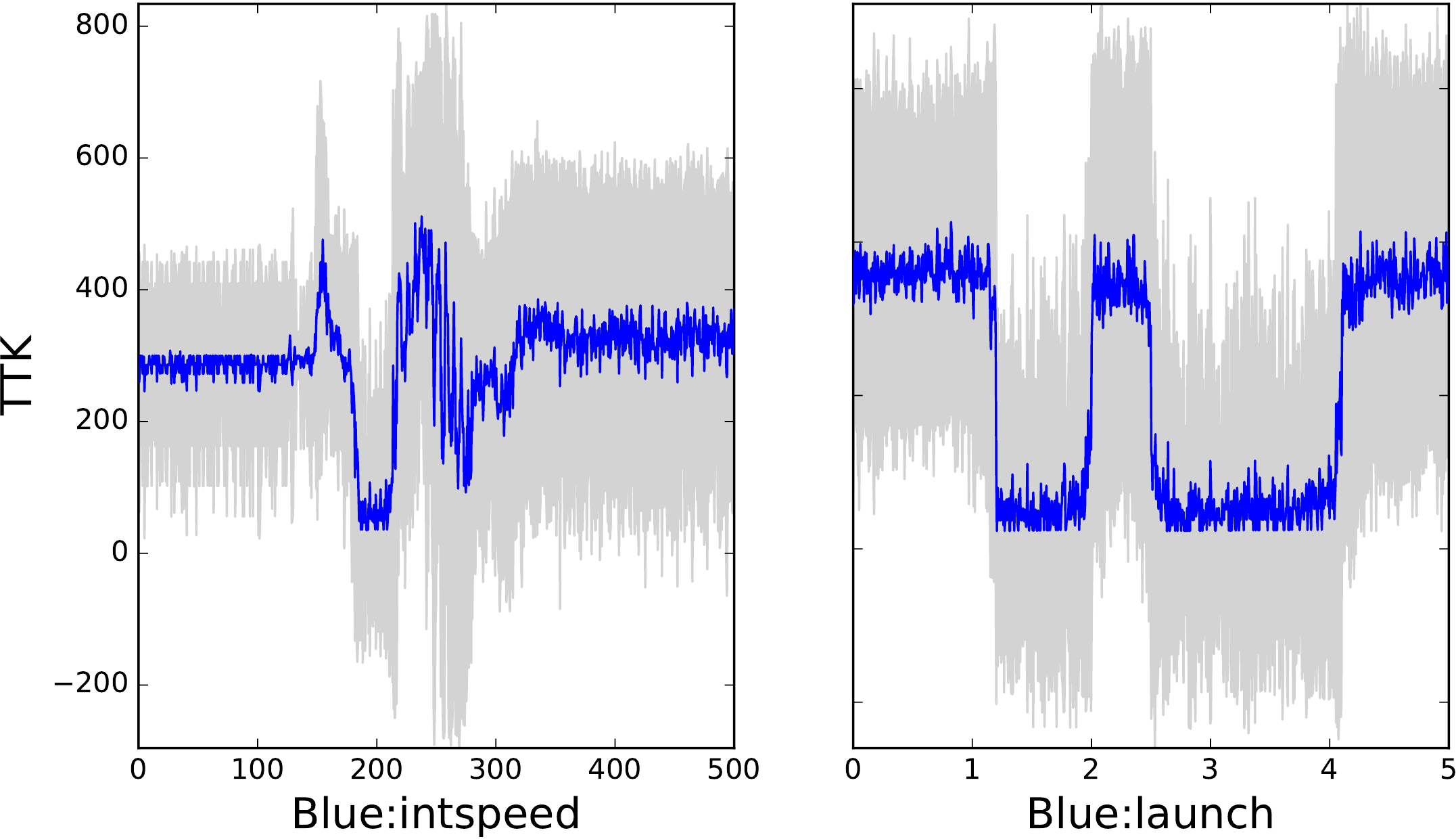}
        \caption{One-dimensional examples of \emph{TTK} objective function, for $\pmb{x}=\{\emph{Blue:intspeed}, \emph{Blue:launch}\}$. These figures were empirically produced by holding all $\pmb{\pmb{x}_r}$ parameters constant, as well as all $\pmb{x}_b$ parameters except the one listed and running many experiments at each location. The dark blue line represents the mean and the shaded area is two times the standard deviation.}
        \label{fig:objective_examples}
    \end{figure}

    In this application, evaluating the objective function (or obtaining the results of a simulated engagement) can be very time consuming. Figure \ref{fig:objective_examples} is an illustration of the TTK objective function based on variations from two different inputs. It is apparent that the TTK function is nonlinear, noisy, and discontinuous. Consequently, a specialized class of optimization is needed that can reduce the number of evaluations needed to find the optimum. We will review, in more detail, the setup of the simulation and how \BO{} was applied.

    \subsection{Learning Engine}
\begin{figure}[htbp]
          \centering
          \includegraphics[width=0.5\textwidth]{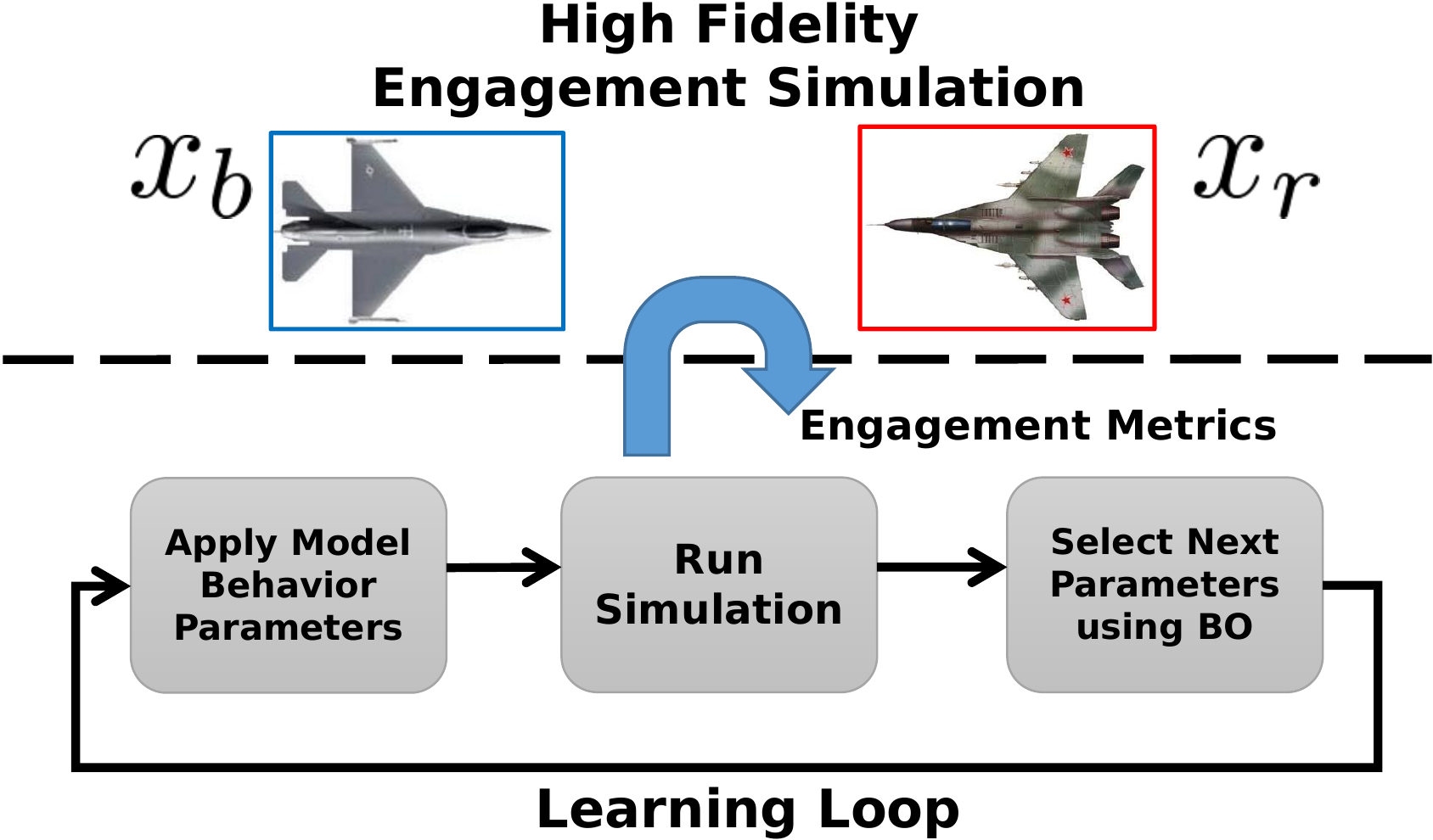}
          \caption{Learning Engine Diagram. Representation of the engagement simulation environment (top) and the high-level learning loop (bottom)}
          \label{fig:LearningEngine}
    \end{figure}

    Figure \ref{fig:LearningEngine} depicts the process by which the optimum parameter set $\pmb{x}_{b}$ is found. The bottom portion of the figure shows the high-level learning loop while the top of the figure represents the simulation environment. Agent parameters are chosen first and then a simulation is run using the parameters for each agent. For this effort, we leveraged pilot agent logic developed by Orbit Logic on previous efforts; however, the overall approach is agnostic of the agent logic itself, such that any agent with tunable behavior parameters could be substituted. After the simulation has completed (based upon defined criteria, such as a time limit), engagement metrics are calculated, and delivered to the  \BO{} algorithm for evaluation. The Bayesian optimization algorithm associates the most recent engagement metrics with the behavior parameters and then selects new parameters with the goal of finding the optimum parameters.

\subsection{Simulation Setup}
    We designed an engagement scenario that included significant opportunity for the collection of relevant metrics to drive and validate the \BO{} learning process. It includes a single blue force fighter jet penetrating an adversary's defended engagement zone. The primary objective is to engage in air-to-air combat against a red defending fighter jet to achieve theater control. The simulation is run for $T_{max}=5$ minutes, or until either the blue or the red agent is eliminated. The blue agent's `goal' is to minimize the TTK, which is defined as the time to the elimination of the enemy. If blue is eliminated the TTK is defined as $2*T_{max}-T_{elim}$ where $T_{elim}$ is the time to when blue was eliminated. Thus, $TTK>300$ represent blue being eliminated, $TTK<300$ represent blue victory, and $TTK=300$ denotes that both fighters survived.

\begin{figure}[htbp]
          \centering
          \includegraphics[width=0.65\textwidth]{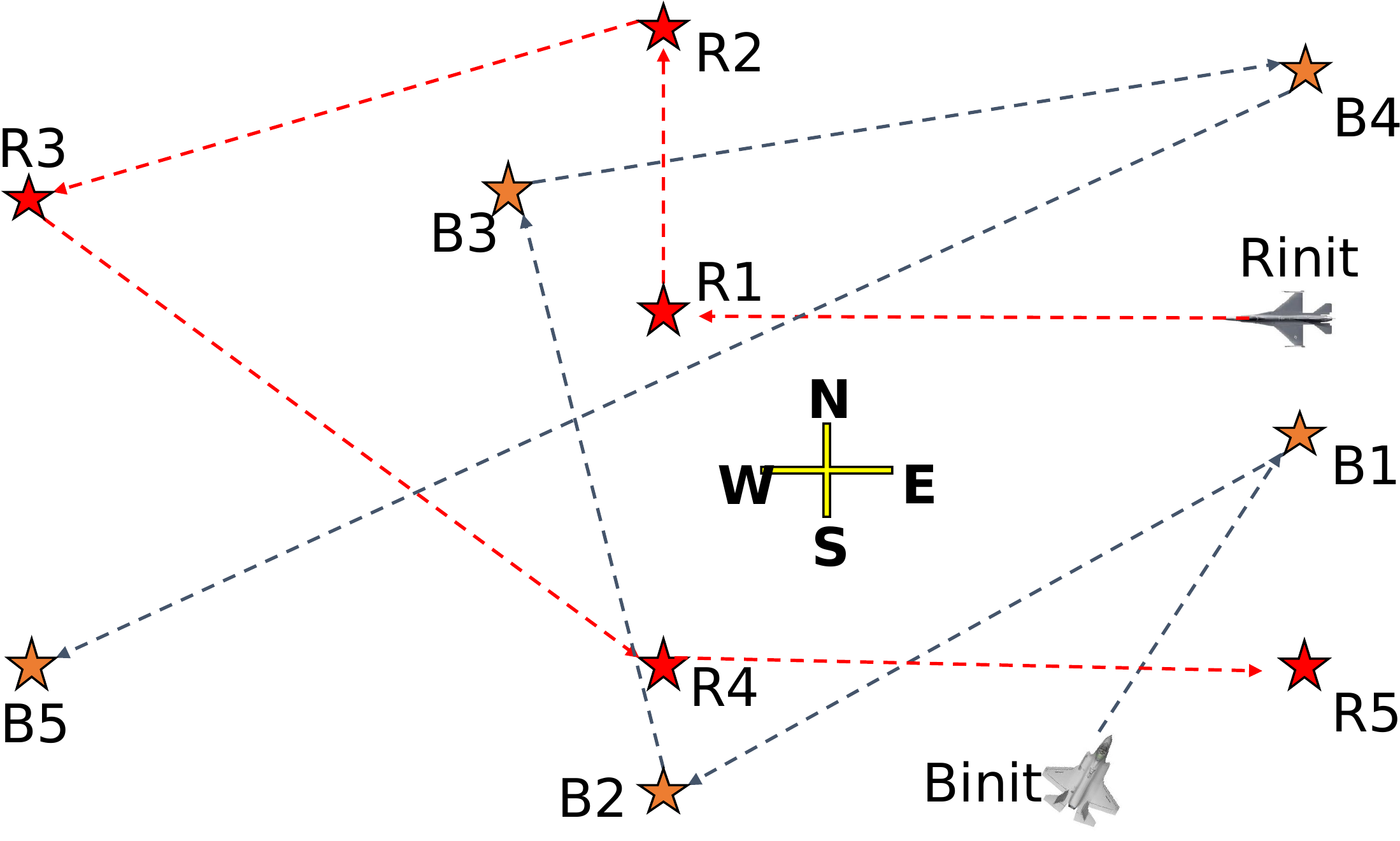}
          \caption{Diagram depicting the engagement space. Red and blue players begin at $R_{init}$ and $B_{init}$ respectively. Their flight path carries them over the way-points indicated by the dashed lines. During the flight they will engage if criteria in their decision logic are met.}
          \label{fig:s1}
    \end{figure}

    Each agent is assigned a nominal flight plan that includes multiple way-points that, when flown over, provide mission `scoring points'. The way-points are arranged in such a way as to ensure that the fighters periodically encounter each other as illustrated in figure \ref{fig:s1}, instigating the employment of air-to-air combat logic. 

    Engagement metrics such as ``TTK'', ``energy management'', or ``number of objectives completed'' are calculated at the conclusion of the simulation and then delivered to the Learning Engine.

    \subsection{Practical Implementation of~\BO{}}\label{practical-application-of-bo}

    While theoretically attractive, na\"{i}ve application of \BO{} will not typically yield satisfactory results. This is due to the fact that the learning the hyperparameters of a GP surrogate model involves non-convex optimization, and the fact that the objective functions are noisy. Furthermore, without any initial data to guide the learning of the GP model, the GP learning process can converge on incorrect results in early stages of the~\BO{} process. Hence, careful design and implementation practices are needed to ensure reliable performance. The following items are the key components that need to be considered:

\subsubsection{Initial seeding of optimization}\label{initial-seeding-of-optimization}
    A subtle but critical point is that in order for \BO{} to work a sufficiently accurate GP is needed in order to approximate the true objective function. More specifically, it is necessary to have a GP that approximates the actual attributes of the objective function. It is therefore necessary to provide `seed' data to bootstrap the GP. To get a good seed sample of the input space we use random sampling~\cite{Bergstra2012}. However, to ensure sufficient coverage of the space, na\"{i}ve uniform random sampling is insufficient because the samples are not correlated and can be drawn in proximity to each other. Instead a type of stratified random sampling called Latin hypercube (LH) sampling~\cite{Macdonald2009,Schonlau1997} is used. This method ensures a more uniform sampling of the input space.

    The number of seed points must also be selected for a given random sampling method. The main factors that would affect how many samples are needed are the homogeneity of the objective function (i.e. whether it behaves similarly over the solution space), and the smoothness of the function. Unfortunately, the objective function is unknown, which limits the ability to optimally select the number of seed points.

    This is important in our work because of the desire to ensure that \BO{} can choose a good experiment location when the optimization process begins. If the sampling is not done appropriately, the \BO{} algorithm can (and likely will) begin investigating a local optimum and not be able to escape. To the best of our knowledge, using some form of quasi random sampling with a heuristically selected amount of seed data is currently state-of-the-art~\cite{Hoffman2014,Chevalier2013}. It is common to use $10d$ seed points, where $d$ is the dimensionality of the problem\cite{Jones1998,Huang2005}. In section \ref{parallel-sampling}, we suggest that, depending on the properties of the objective function, it is necessary to use a different heuristic other than a simple multiple of the dimensionality.

\subsubsection{Hyperparameter Learning}\label{MLE_MAP_CV}
    As the true kernel hyperparameters $\Theta$ are not known a priori it is extremely important to be able to learn them in the most accurate and reliable way possible. The hyperparameters are the parameters that govern how the covariance function (or kernel) behaves (see Equation~\ref{eq:Mat32}). As more data is used in the kernel, the hyperparameters can be trained to best fit the data.

    There are several approaches to learning the hyperparameters. Maximum Likelihood Estimate (MLE), Cross-Validation (CV), and Maximum a Posteriori (MAP) are three widely used methods that we investigated. Each of these methods yields a point estimate of the `true' hyperparameters, which can be used in the GP kernel.

    In our learning engine implementation the MAP estimate is used. This method was found to be the most stable and tolerant of different objective functions and simulation settings. Both MLE and cross-validation were very sensitive and tended to over-fit the training data\footnote{Because there is very little data, due to the fact that function evaluations are to be minimized}. This is undesirable, because if the hyperparameters are selected poorly (yielding a GP surrogate function that does not reflect the true objective function ``well enough'') the~\BO{} will begin searching in the wrong places. Despite the use of MAP, there are still some conditions presenting stability difficulties. Future efforts will investigate Monte-Carlo methods and use distributions over the hyperparameters instead of explicitly choosing the best estimate~\cite{Snoek2012}.

\subsubsection{Parallel Sampling}\label{parallel-sampling}
    Using an acquisition function $a(\cdot)$, there are two commonly used ways to search for the optimum. The first is single sampling (SS) where $y_{i}(\pmb{x})$ is evaluated a single time at the $\argmax_{\pmb{x}}a(\cdot)$ of the acquisition function; this is the standard \BO{} approach. The second is multiple (or batch) sampling (MS), in which $y_{i}(\pmb{x})$ is evaluated at multiple different locations simultaneously. Finally, we propose a method called repeat sampling (RS), which is identical to SS except that the objective function will be evaluated repeatedly at the same location, $\argmax_{\pmb{x}}a(\cdot)$.
        
    The intuitive reason for introducing repeat sampling is to obtain a more informative statistical sample of the objective function at every iteration. This is necessary because, for \BO{} to work properly, the surrogate function needs to be a `sufficiently accurate' representation of the true objective function. RS helps the GP to have more information regarding the noise of the true objective function, so that the GP can be a useful surrogate function in guiding the \BO. The concept of repeat sampling is also commonly used in experimental design where is is known as experiment ,eplication \cite{pyzdek2003quality,croarkin2006nist}. Besides gaining more statistical information, replication is frequently used when experiment setup is expensive; this would definitely be a consideration when training against human pilots.

    \paragraph{Multi-point Sampling}
    Several methods can be used for MS~\cite{Snoek2012,Chevalier2013,Contal2013}. Generally, these methods attempt to forecast which points might be of the most interest in future evaluations. Three different methods are investigated in this paper. 
            
    The first is known as $q$-EI~\cite{Schonlau1997,Ginsbourger2010}. It is similar to EI but instead of optimizing over $\pmb{x_b}$ the EI function is optimized over multiple ($q$) $\pmb{x}_b$ points at the same time. This method is notoriously expensive to evaluate, not only because the dimensionality increases (e.g. with $q=5$ and $d=2$ $q$-EI would involve optimizing 10 variables instead of 2) but because the calculation of the EI function is more expensive. It is so expensive that Monte-Carlo simulations are known to be faster in some cases. As an alternative an approximate method was introduced for computing the $q$ points more quickly~\cite{Chevalier2013}. This method is used in our simulations.

    The next method called GP-UCB-PE uses the UCB function and a pure exploration (PE) technique to perform parallel search for the optimum~\cite{Contal2013}. The GP-UCB-PE algorithm has a simple premise: greedily select the first $q$ (borrowing notation from $q$-EI) points using the UCB acquisition function. This is done by first selecting the maximum of the UCB function as $x_1$, identical to the non-batch approach. Next, taking advantage of the fact that the covariance predictions can be updated without knowing $y(x_1)$, the UCB can be re-calculated assuming the proposed $x_1$ were known and select the resulting maximum as $x_2$. The process is repeated until $q$ points have been selected.

    Finally, a simple extension of basic TS can be used to select MS points. Instead of drawing a single function and selecting its optimum, $q$ functions are drawn, the optimum of each is found, and experiments are run at each of the corresponding locations of an optimum. TS is known to have difficulty in high-dimensions because of the curse of dimensionality. Drawing random functions from high-dimensional space is not easy because, in order to get sufficient resolution, one must randomly sample over the entire input space $x_b$.

    \paragraph{Repeat Sampling}
    RS is critical in problems with extremely noisy and volatile objective functions (like those found in simulated aerial combat). It allows the GP to converge for a noisy objective function (returning to the point made in subsection \ref{initial-seeding-of-optimization}). Intuitively, this can be understood by recognizing how Algorithm \ref{alg:BayesOpt} works, that is: the kernel is updated at the end of each iteration. If only a single sample is evaluated in an iteration then the kernel update could be responding to noise and the acquisition function could then direct the next experiment to the wrong place. If instead the kernel is provided with a `statistical sample' of the function the update will be less likely to respond to outlying observations.

    \paragraph{Hybrid Repeat/Multi-point Sampling}
    The RS and MS strategies are especially valuable when the objective function is less expensive to evaluate via simulation, and the experiments can be run in parallel without significantly increasing the overall cost of the optimization. In the following, MS=3 means that 3 batch samples will be selected. Likewise, RS=3 is where 3 samples will be taken at the same location. Note that SS is a special case where RS=MS=1. Finally, we refer to combined sampling strategies using both RS and MS as Hybrid Repeat/Multi-point Sampling (HRMS). HRMS will be experimentally evaluated in section \ref{sec:results}. The goal will be to identify how varying RS and MS affects \BO.

\section{Results}\label{sec:results}
    Given a decision agent optimization problem, we perform experiments to investigate the performance of different acquisition functions for the aerial combat simulations. More importantly we wish to investigate the effect that varying RS and MS has on the optimization results.

    Specifically, three common acquisition functions will be evaluated: Expected Improvement (EI)~\cite{Jones1998}, upper confidence bound (GP-UCB)~\cite{Srinivas2010}, and Thompson Sampling (TS)~\cite{Thompson1933}. We also evaluate their corresponding batch sampling forms: q-EI~\cite{Wang2016}, GP-UCB-PE ~\cite{Contal2013}, and multiple draws from TS. The different levels of RS and MS used are $RS=\{1,3,5,10\}$ and $MS=\{1,3,5\}$. The GPML toolbox is used for GP representation and hyperparameter inference~\cite{Rasmussen2006b}. Orbit Logic developed discoverable data interfaces usable to set blue and red force agent parameters, initiate simulation runs, and recover run metrics at the conclusion of each. Other relevant experiment parameters are shown in Table~\ref{tab:experiments}.
    
    \begin{table}[]
        \centering
        \caption{Experiment Configuration}
        \label{tab:experiments}
        \begin{tabular}{@{}cc@{}}
        \toprule
        Property                               & Configuration                       \\
        \midrule
        \multicolumn{1}{l|}{Acq. Fxns.}        & EI, UCB, TS                          \\
        \multicolumn{1}{l|}{$\pmb{x}$}               & $\{x1,x2\} = \{launch,intspeed\}$    \\
        \multicolumn{1}{l|}{$y$}               & $ TTK$                         \\
        \multicolumn{1}{l|}{\# of Seeds}       & $10d=20$                              \\
        \multicolumn{1}{l|}{Kernel}            & M\'{a}tern ARD, $\nu=3/2$            \\
        \multicolumn{1}{l|}{Mean Hyperpriors}  & $\mathcal{N}(\mu=0,\sigma^2=100^2)$  \\
        \multicolumn{1}{l|}{Cov. Hyperpriors}  & $\mathcal{U}(\log 1, \log 3)$                   \\
        \multicolumn{1}{l|}{Lik. Hyperprior}   & $\mathcal{U}(\log 20, \log 400)$                 \\
        \multicolumn{1}{l|}{Inference}         & Laplace                              \\
        \multicolumn{1}{l|}{Experiments per Condition}            & 4                 \\
        \midrule
        \end{tabular}
    \end{table}

    In Table~\ref{tab:experiments} the acronym ARD stands for Automatic Relevance Determination. ARD involves modifying the kernel so each input has its own scaling. ARD is particularly useful for dealing with mixed engineering units, or when there may be inputs that are not as `relevant' in contributing to the output as others. The hyperparameters are estimated using MAP. MAP requires that prior probabilities (or beliefs) be placed on each hyperparameter. The prior beliefs on hyperparameters are known as hyperpriors. Finally, when using MAP with non-Gaussian hyperpriors the posterior distribution $p(y|X)$ will not be an analytical distribution and Laplace inference is used as an approximate method for finding the maximum of the posterior distribution.

\begin{figure}[htbp]
        \centering
        \includegraphics[width=0.90\textwidth]{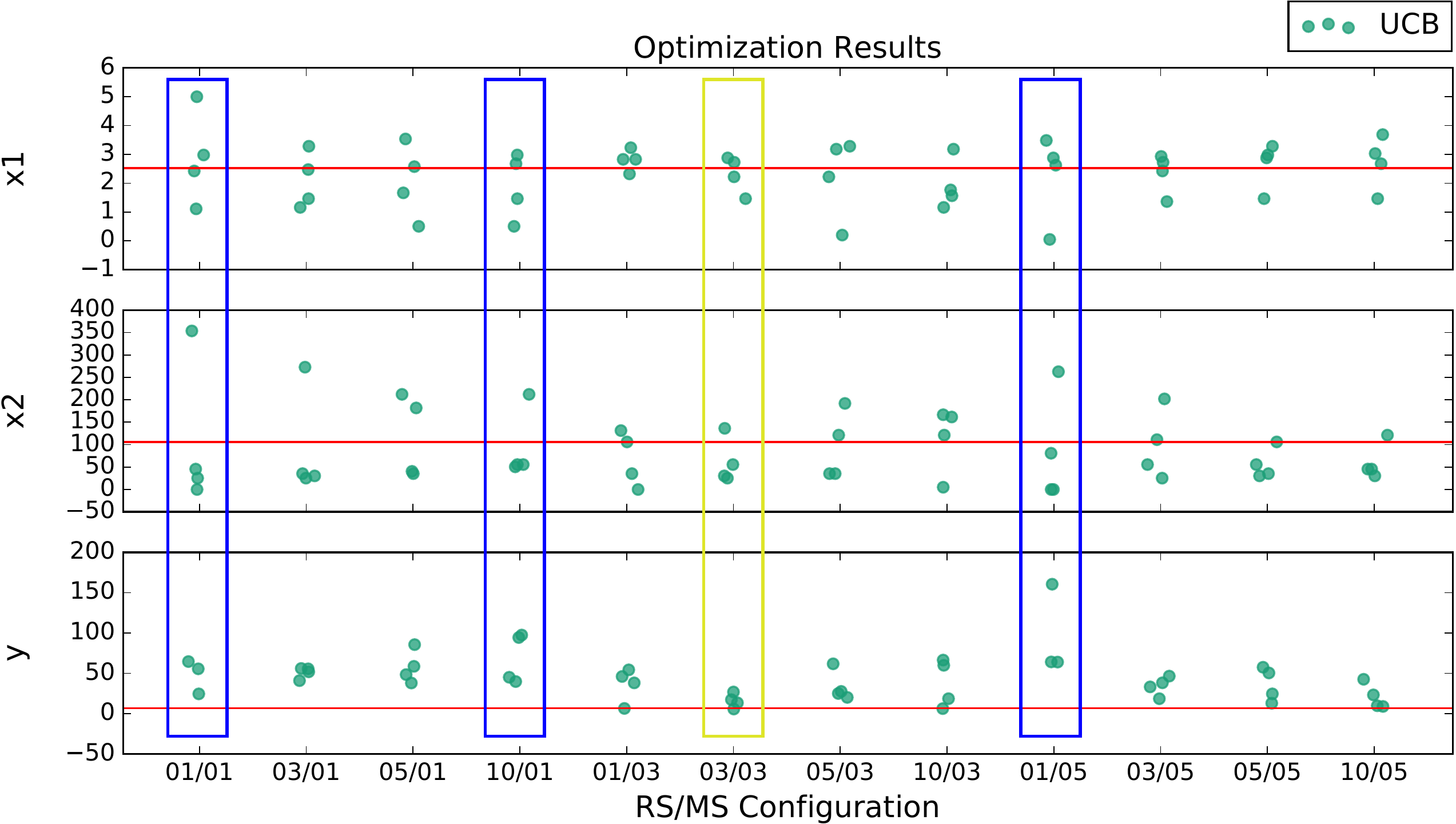}
        \caption{Scatter plots of the $x1,x2$ and $y$ values for different RS/MS configurations. Results from running \BO{} for approximately 500 function evaluations. The red horizontal line is the ground truth value.}
        \label{fig:2d_time_plot_comparison}
    \end{figure}

    Figure~\ref{fig:2d_time_plot_comparison} shows the estimated locations of the optimum as well as the optimum itself for the UCB acquisition function. The estimates for $\pmb{x}$ and $y$ grow tighter together, and closer to the ground truth, as both RS and MS become greater than 1. The configurations marked by colored rectangles highlight that methods using solely SS, RS, and MS (shown in blue), underperform the method that combines both RS and MS greater than one (yellow). This finding is similar for the EI and TS functions as well. From this figure we can conclude that there are HRMS configurations that yield more repeatable optimization results than SS, and that neither RS or MS alone is clearly better.

    On more detailed investigation of the results, for large RS the optimization tends to terminate early due to an ill-conditioned covariance matrix. This occurred because RS is `too big' at those locations, having returned too many nearly identical objective function values at the same $(x_1,x_2)$ locations. The subsequent covariance matrix became too linearly dependent, which led to conditioning problems for GP inference. This suggests that there is clearly a trade-off between the benefits of RS and an unstable GP. We revisit this point in the conclusion section.

\begin{figure}[htbp]
        \centering
        \includegraphics[width=0.90\textwidth]{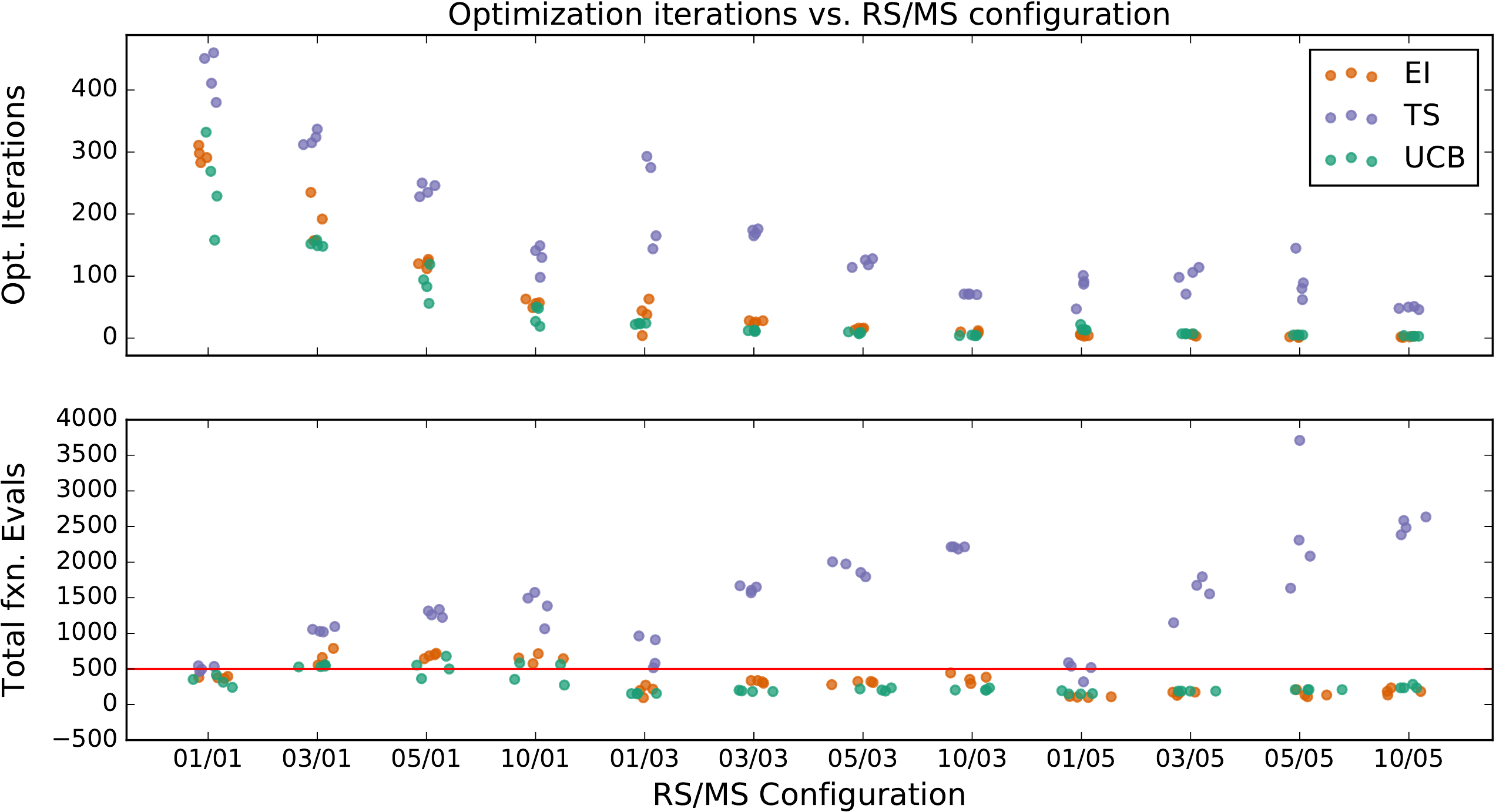}
        \caption{ Plot showing the total amount of optimization iterations (Top), and the corresponding number of function evaluations (Bottom) after running each configuration for 2 hours. Note that, with the exception of TS, the total number of function evaluations for mixed RS/MS configurations generally doesn't exceed that of \BO{} with SS}
        \label{fig:tot_evaluations}
    \end{figure}

    It is important to note that given a fixed time for optimization (i.e. not limiting the function evaluations for methods that perform more quickly), the total number of function evaluations in most cases does not exceed that of the SS strategy. This is mainly due to the overhead of calculating MS, and is illustrated in Figure~\ref{fig:tot_evaluations}, where only TS significantly exceeds the total function evaluations of SS. 

\begin{figure}[htbp]
        \centering
        \includegraphics[width=0.90\textwidth]{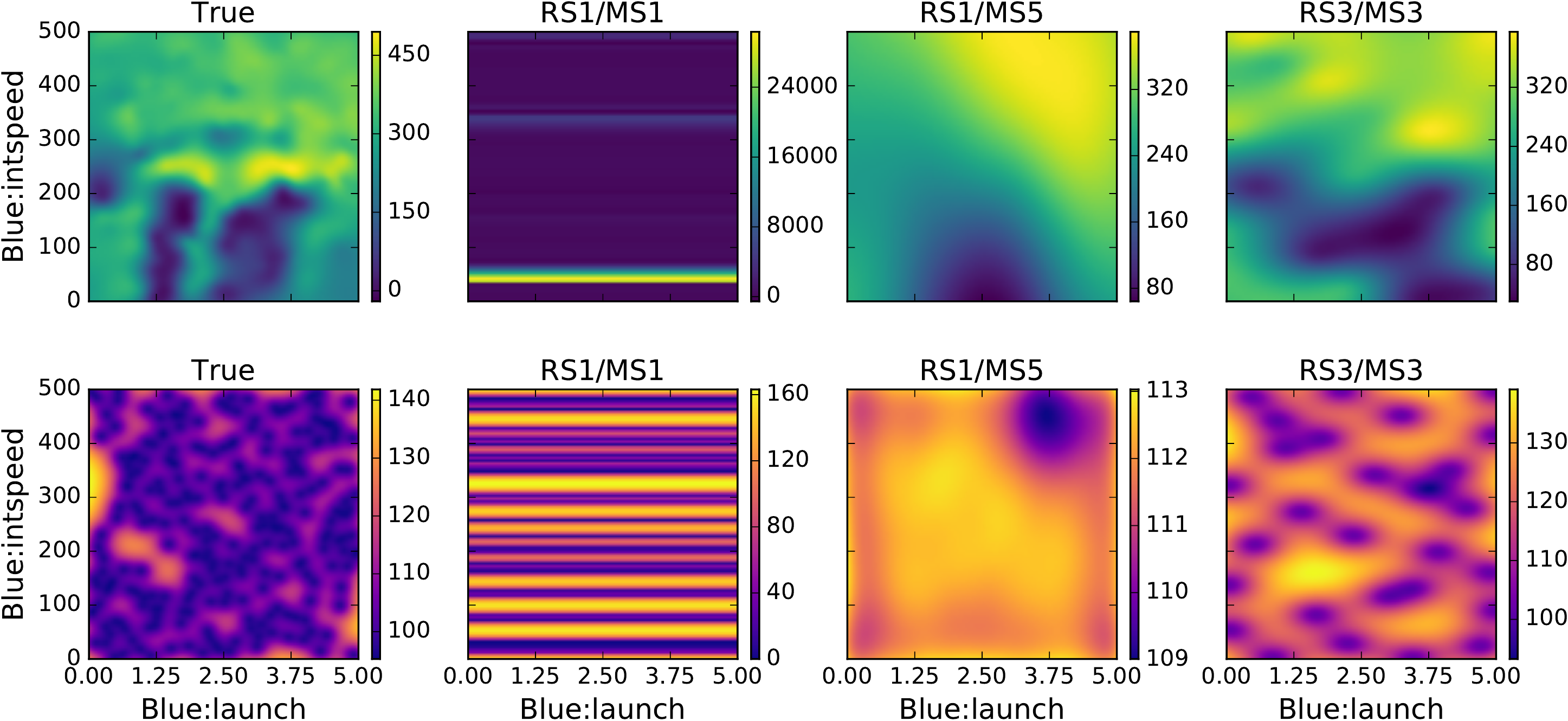}
        \caption{Table of figures illustrating the effect of combined RS/MS sampling using the UCB acquisition function. Top row is $\mu_{GP}$ bottom row is $\sigma_{GP}$. From left to right the first column is the truth surface obtained by high density sampling and fitting a GP to the data. The following columns show some example results from optimization runs using the indicated values for RS and MS. Each of the final 3 columns represents the optimization solution after 2 hours}
        \label{fig:sample_instability}
    \end{figure}

    Figure~\ref{fig:sample_instability} depicts some examples of the final GPs obtained during time limited optimization (again, allowing faster methods to use more evaluations/iterations) for three different HRMS configurations. The far left column is the `ground truth GP' model that is obtained by training with several thousands of samples over the input space. The key insight is that the RS3/MS3 strategy yielded a GP that better represents the ground truth. 
    
    These findings indicate that HRMS both improves the repeatability of the optimization, \emph{and} the overall fidelity of the surrogate representation of the objective function. This applies for both a fixed computation time (i.e. not limiting faster methods like SS to have the same number of function evaluations), and number of function evaluations. These two phenomena are linked, i.e. the optimization is more repeatable (and reliable) because the surrogate representation is more accurate.

    Returning to the high level goal of this research: training an AI with behavioral parameters to optimize a combat objective, and to have the capability of being adaptive. Figure~\ref{fig:2d_time_plot_comparison} demonstrates that \BO{} is useful in this application. It shows that using HRMS results in more repeatable identification of the optimum AI behaviors in the simulated air combat.
    
\begin{figure}[htbp]
        \centering
        \includegraphics[width=0.90\textwidth]{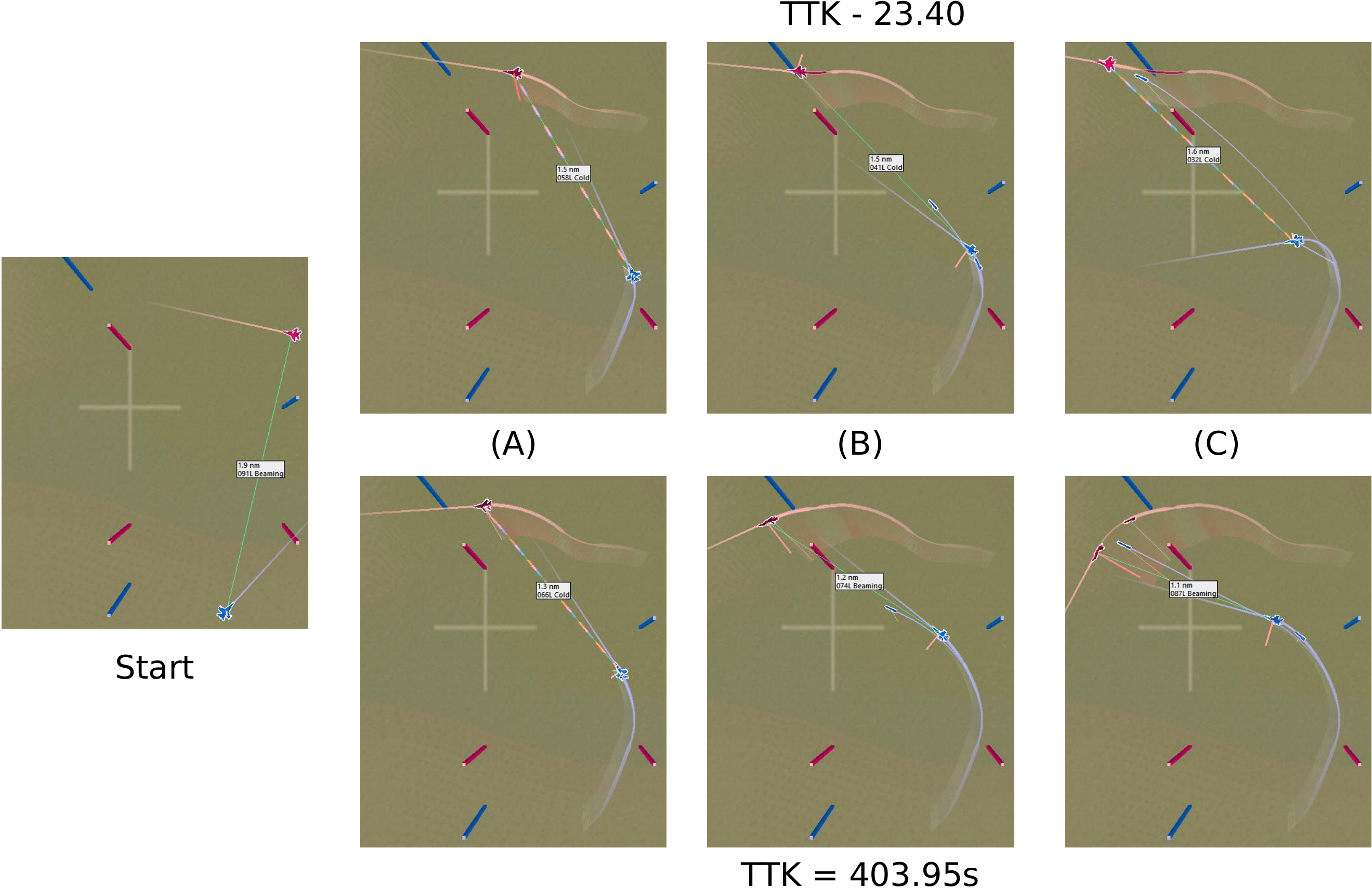}
        \caption{Comparison between $x_b=\{2.60,125.0\}$ with associated $y=23.4$ (top) and $x_b=\{3.00,450.0\}$ with $y=403.95$ (bottom). The far left image shows the simulation start which is identical for both scenarios. The images in the (A) column show the point at which the blue agent launches a mid-ranged missile. The images in column (B) show that in both scenarios blue loses lock on red. However, in column (C) the blue with lower intercept speed is able to regain lock, while on the bottom blue is not able to maneuver in order to regain lock in time, without lock the missile is not longer guided to it's target.}
        \label{fig:tacview}
    \end{figure}

    Figure~\ref{fig:sample_instability} shows that the combined RS3/MS3 strategy (column 4) yields a more accurate surrogate representation of the true objective function. This surrogate model gives insights about how the expected value of the TTK changes over the entire behavioral space. There seem to be two large areas with lower $intspeed$ and moderate $launch$ that are good for blue. Conversely it appears that the higher $intspeed$ configurations don't yield very good results. Looking at the simulation recording helps gather more insight as to why. Figure \ref{fig:tacview} shows the outcomes of two different simulations. A scenario with lower $launch$ and $intspeed$ values that has good results, and a scenario with higher $intspeed$ that yields poor results because the blue agent cannot regain lock after launching a mid-range missile. In this simulation the jet must have an active lock on the target to continue guiding the missile.

    Several other interesting insights can be drawn from the surrogate identified with RS3/MS3, while the surrogate functions identified with SS and to a lesser extent RS1/MS5 (columns 2 and 3) offer less reliable insight into the true TTK function. This capability gives the AI the ability to identify the optimum behavior, but also have a better capability to adapt by being able to predict outcomes at any other location in the behavioral space.

\section{Conclusion}
    We have demonstrated the ability to tune the behavior parameters of an autonomous agent. In this case we were able to select the behavioral parameters of the blue AI pilot in order to optimize its TTK. Beyond optimizing the TTK we have shown that it is possible to construct a useful surrogate model during the optimization process. \BO{} with HRMS sampling was used in order to achieve these goals, using HRMS together with \BO{} is necessary due to the characteristics of the of simulated combat problem, the objective function is extremely volatile and it is desirable to not only learn an optimum action but to model the objective function as well. 

    \BO{} provides a way to perform optimization on expensive objective functions. This is due to the ability to make predictions about the value of the objective function in locations that have not yet been investigated. Using HRMS can improve this ability. As such, both the final optimization results are improved, and valuable insights about the true objective function are modeled. An AI with a better global model of the true objective is better able to adapt in order to achieve desired outcomes.

    Using \BO{} with HRMS has several advantages in this application. As demonstrated, the algorithm was able to identify the optimum in extremely nonlinear and noisy objective functions. Just as importantly, each evaluation of the objective function was specifically chosen to yield the most benefit with the given data, which kept the number of combat simulations (function evaluations) as low as possible.

    Of course, we only used simulated engagements between two AI's, applying this methodology to simulations with human participants is required to really address the higher-level goals of this research. It would also be interesting to see how decision-making AI could use the model of the objective function not only to be able to select behaviors in a noisy environment, but to be able to act as a tutor to the human, suggesting behaviors that will most likely yield successful results.

    Regarding application to decision-making AI, we wish to re-emphasize that the concepts applied here can be be extended to simultaneously train multiple red and blue force agents, or to train team behaviors as well. Also, With the addition of statistical behavior recognition components and exploitation of the ability to run \BO{} with parallelizable simulations, the fundamental learning approach could also be quite naturally applied to extremely rapid online learning and adaptation, as well as learning with humans in the loop.

    With respect to the theory of \BO{} with HRMS sampling, further work is needed to investigate autonomous methods in which the values for RS and MS can be calculated to attempt to guarantee the best possible performance. The existing literature regarding experiment replication may be helpful in this regard. It may also be of importance to adaptively adjust RS due to problems with it possibly being `too big' such that ill-conditioned covariances are introduced, as mentioned above. This can be of particular concern with a heteroskedastic objective function, or in other words an objective function where the statistical properties of the noise varies over the input space.

    A couple of theoretical challenges still present themselves and should be investigated in future work. The first is that the hyperparameters remain sensitive even with the improvements from using MAP, instead using fully Bayesian learning (where the hyperparameters are marginalized out of the problem) might be promising. However, this approach involves several of its own drawbacks such as requiring the use of MCMC sampling. The second challenge is that the curse of dimensionality is ever present. The methods by which GP-UCB-PE and TS are calculated do not transfer well to higher dimensions (basically requiring a large number of samples randomly selected over the entire solution space). Instead, it would be useful to utilize the output of the DIRECT optimization algorithm (something that is already used and that can have a fixed cost) to indicate where the most important locations to sample the objective function are located in the high dimensional space, and to sample more densely there. Also, a method using ``Random Embeddings'' has been introduced to enable Bayesian optimization to be useful in problems with much higher dimensionality, but remains to be evaluated in this setting~\cite{Wang2013}.

\bibliography{References}
\end{document}